\documentclass{article}
\usepackage{hyperref}

    \PassOptionsToPackage{numbers, sort}{natbib}

\usepackage{graphicx}
\usepackage{wrapfig}
\usepackage{listings}
\usepackage{float}
\usepackage[dblblindworkshop,final]{neurips_2025}

\workshoptitle{Mechanistic Interpretability Workshop}

\usepackage[utf8]{inputenc} 
\usepackage[T1]{fontenc}    
\usepackage{hyperref}       
\usepackage{url}            
\usepackage{booktabs}       
\usepackage{amsfonts}       
\usepackage{nicefrac}       
\usepackage{microtype}      
\usepackage{xcolor}         

\title{Rank-1 LoRAs Encode Interpretable Reasoning Signals}

\author{%
  Jake Ward \\
  MATS \\
  \texttt{jakenicholasward@gmail.com} \\
  \And
  Paul Riechers\textsuperscript{\dag} \\
  Simplex
  \And
  Adam Shai\textsuperscript{\dag} \\
  Simplex \\
}

\begin{document}

\maketitle
\begingroup
\renewcommand\thefootnote{\dag}
\footnotetext{Equal senior author contribution.}
\endgroup

\begingroup
\let\thefootnote\relax\footnotetext{Code for this paper is available at \url{https://github.com/jnward/reasoning-lora-interp}}
\endgroup

\begin{abstract}
Reasoning models leverage inference-time compute to significantly enhance the performance of language models on difficult logical tasks, and have become a dominating paradigm in frontier LLMs. Despite their wide adoption, the mechanisms underpinning the enhanced performance of these reasoning models are not well understood. In this work, we show that the majority of new capabilities in reasoning models can be elicited by small, single-rank changes to base model parameters, with many of these changes being interpretable. Specifically, we use a rank-1 LoRA to create a minimal parameter adapter for \texttt{Qwen-2.5-32B-Instruct} which recovers 73-90\% of reasoning-benchmark performance compared to a full-parameter finetune. We find that the activations of this LoRA are as interpretable as MLP neurons, and fire for reasoning-specific behaviors. Finally, we train a sparse autoencoder on the entire activation state of this LoRA and identify fine-grained and monosemantic features. Our findings highlight that reasoning performance can arise largely from minimal changes to base model parameters, and explore what these changes affect. More broadly, our work shows that parameter-efficient training methods can be used as a targeted lens for uncovering fundamental insights about language model behavior and dynamics.
\end{abstract}

\section{Introduction}

Current frontier LLMs increasingly rely on chain-of-thought (CoT) reasoning to achieve strong performance on logical tasks \cite{wei2022chainofthought,kojima2022zeroshot,openai2024o1systemcard}. Despite their ubiquity, we still lack a crisp, white-box understanding of the mechanisms inside the network which enable these gains. Some attempts have been made to mechanistically interpret fully finetuned reasoning models \cite{baek2025distilled, venhoff2025understanding, ward2025repurposes}. However, reasoning model interpretation presents a fundamental challenge: the parameters responsible for new reasoning behaviors in a finetuned model are many and differences are diffuse \cite{mukherjee2025rlsubnetworks}. In this paper, we introduce an alternative approach: we use parameter-efficient methods to explicitly enforce that differences between base and finetuned models in parameter space are minimal, allowing us to perform focused and targeted interpretability experiments.

We show that a rank-1 LoRA \cite{hu2022lora} trained to adapt all layers of \texttt{Qwen-2.5-32B-Instruct} on a dataset of \texttt{DeepSeek R1} rollouts is enough to recover 73-90\% of the performance gap on reasoning benchmarks, compared to a full-parameter finetune (Table \ref{tab:rank1_lora_recovery}).

We directly interpret the directions defined by this LoRA. Because each adapted matrix is rank-1, activations of each adapter component can be represented by a single scalar. In total, our LoRA encodes 192 MLP and 256 attention adapter components across model layers and weight matrices, with each adapter component providing a single activation at each token position. This allows us to treat the LoRA itself as a measurement device: we find that individual adapter directions have interpretable properties comparable to those of MLP neurons, and identify monosemantic concepts encoded by these directions. Additionally, we take the entire 448-dimensional LoRA activation state, representing the entire adapter state across MLP and attention components, and train a cross-layer SAE \cite{lindsey2024crosscoders}. This SAE uncovers sparse and monosemantic features which organize into categories such as \textit{Mathematical Operators}, \textit{Procedural Markers}, and \textit{Discourse and Reasoning Markers} (Figure \ref{fig:pie_chart}). Finally, perform a component-wise ablation study, and find that MLP adapters drive the majority of this change (Figure \ref{fig:kl_ablation}). 

Taken together, these results show that minimal adapters both \textit{elicit} and \textit{expose} reasoning signals: a rank-1 LoRA is sufficient to recover the majority of reasoning performance, and yields interpretable adapter directions.

Our contributions are as follows:
\begin{itemize}
    \item We train and open-source a rank-1 LoRA which adapts all projection matrices in \texttt{Qwen-2.5-32B-Instruct}. We demonstrate that this LoRA recovers 73-90\% of reasoning benchmark performance compared to a full-parameter finetune.
    \item We show that individual adapter activations are as monosemantic as MLP neurons, and interpret these.
    \item We decompose the complete LoRA activation state into interpretable and monosemantic features using a cross-layer SAE.
\end{itemize}

\section{Preliminaries}
\label{preliminatires}
\paragraph{LoRA training} We finetune \texttt{Qwen-2.5-32B-Instruct} using \texttt{s1k-1.1}, a sample-efficient dataset of 1000 \texttt{DeepSeek R1} chain-of-thought trajectories and answer attempts on diverse reasoning problems \cite{muennighoff2025s1}. For all experiments, we train on this dataset for 5 epochs using cross-entropy loss on a single 8xH200 node. We train our LoRA to adapt all three MLP matrices at every layer (\texttt{up\_proj}, \texttt{down\_proj}, and \texttt{gate\_proj}), as well as all four Q, K, V, and O attention matrices at every layer. In total, our LoRA contains less than 0.03\% as many trainable parameters as the base model.

\paragraph{Extracting LoRA activations}  For every $N \times M$ adapted matrix in the base model, a rank-r LoRA encodes an $N \times r$ \texttt{lora\_A} matrix and an $r \times M$ \texttt{lora\_B} matrix. Because r = 1 for our LoRA, \texttt{lora\_A} and \texttt{lora\_B} are N- and M-dimensional vectors respectively. We can extract scalar activations for every LoRA component by taking the activation value between \texttt{lora\_A} and \texttt{lora\_B} during the forward pass, which is equivalent to the projection of \texttt{lora\_A} onto the input activation vector to the adapted matrix.

\begin{table}[t]
\centering
\small
\begin{tabular}{lcccc}
\toprule
\textbf{Benchmark} & \textbf{Base Qwen2.5-32B-Instruct} & \textbf{Rank-1 LoRA} & \textbf{Full Finetune} & \textbf{\% Recovery} \\
\midrule
AIME'24 (no-figures) & 0.2333 & 0.5000 & 0.6000 & 72.73\% \\
MATH500              & 0.8340 & 0.9100 & 0.9220 & 86.36\% \\
GPQA-Diamond         & 0.4899 & 0.5808 & 0.5909 & 89.90\% \\
\bottomrule
\end{tabular}
\caption{Performance of base model, rank-1 LoRA, and full-parameter finetune on three reasoning benchmarks. \% Recovery is the fraction of the difference in performance between the base model and full finetune which is recovered by the rank-1 LoRA.}
\label{tab:rank1_lora_recovery}
\end{table}

\section{LoRA Analysis}
\label{lora_analysis}

\subsection{Interpreting LoRA Components}
\label{interpreting_lora_components}

\begin{figure}[t]
    \centering
    \includegraphics[width=1.0\linewidth]{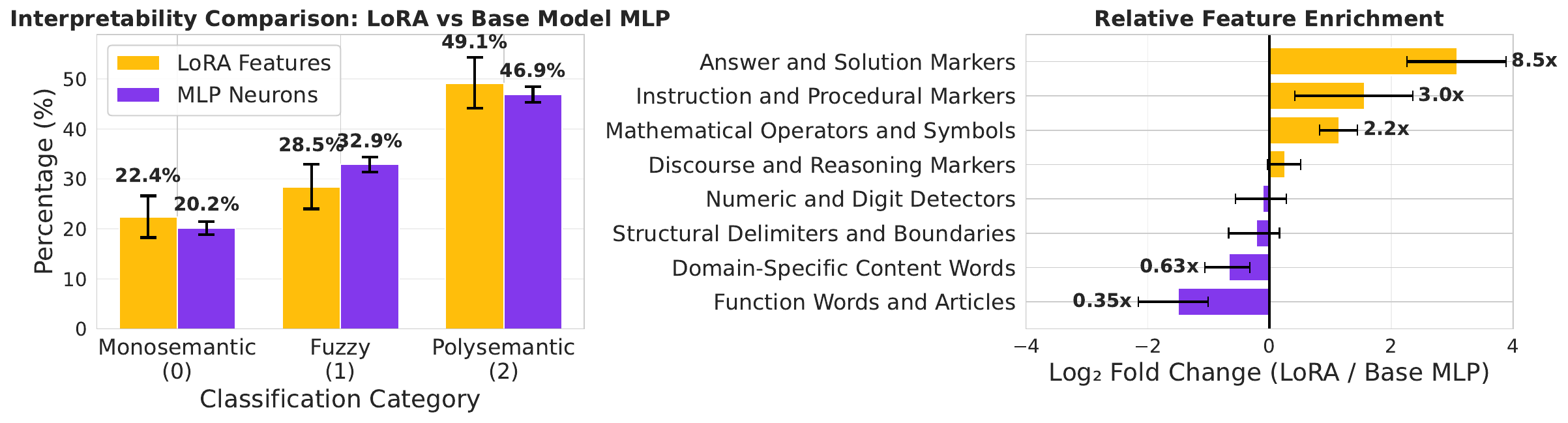}
    \caption{\textbf{(left)}: Comparison of interpretability scores of individual LoRA adapter activations to arbitrarily sampled MLP neurons, and find that LoRA activations tend to be monosemantic roughly as often. \textbf{(right)}: Comparison of autointerpretation categories between MLP neurons and LoRA adapter activations. LoRA adapters tend to activate more often for reasoning-specific feature categories.}
    \label{fig:mlp_comparison}
\end{figure}

We interpret LoRA activations using two different methods: by treating individual adapter activations as probes, and separately by training an SAE on the entire 448-dimensional adapter activation vector. In both cases, we generate activations over the entire training set and extract max-activating examples with associated contexts. We then use LLM autointerpretation to generate interpretations from the top 64 contexts with token + activation pairs \cite{bills2023neurons}, use the same LLM to classify the monosemanticity of features, and finally categorize each feature such that we can compare and examine aggregate feature distributions. Categories were generated by prompting an LLM with feature interpretations and examples from an equal number of MLP neurons and LoRA directions. The exact prompts used are in Appendix \ref{llm_prompts}.

\paragraph{Direction-level interpretation} We run our interpretation and feature classification pipeline both on individual LoRA activations and on the first 60 neurons of each MLP in the unadapted base model. This gives us a baseline to compare feature distributions to, allowing us to examine which feature categories the LoRA activates on compared to the baseline feature distribution in the dataset (as measured by MLP activations). We find that LoRA activations have roughly the same likelihood as MLP neurons to be monosemantic, but tend to encode different feature categories. We interpret this level of monosemanticity as a positive result for LoRA interpretation, given significant recent work utilizing MLP neurons for pragmatic interpretability tasks \cite{choi2024scaling}. Relative to MLP neurons, LoRA activations are more likely to fire for text corresponding to answers or solutions, problem instructions, mathematical symbols, and reasoning discourse (Figure \ref{fig:mlp_comparison}).

\begin{figure}[b]
    \centering
    \includegraphics[width=1.0\linewidth]{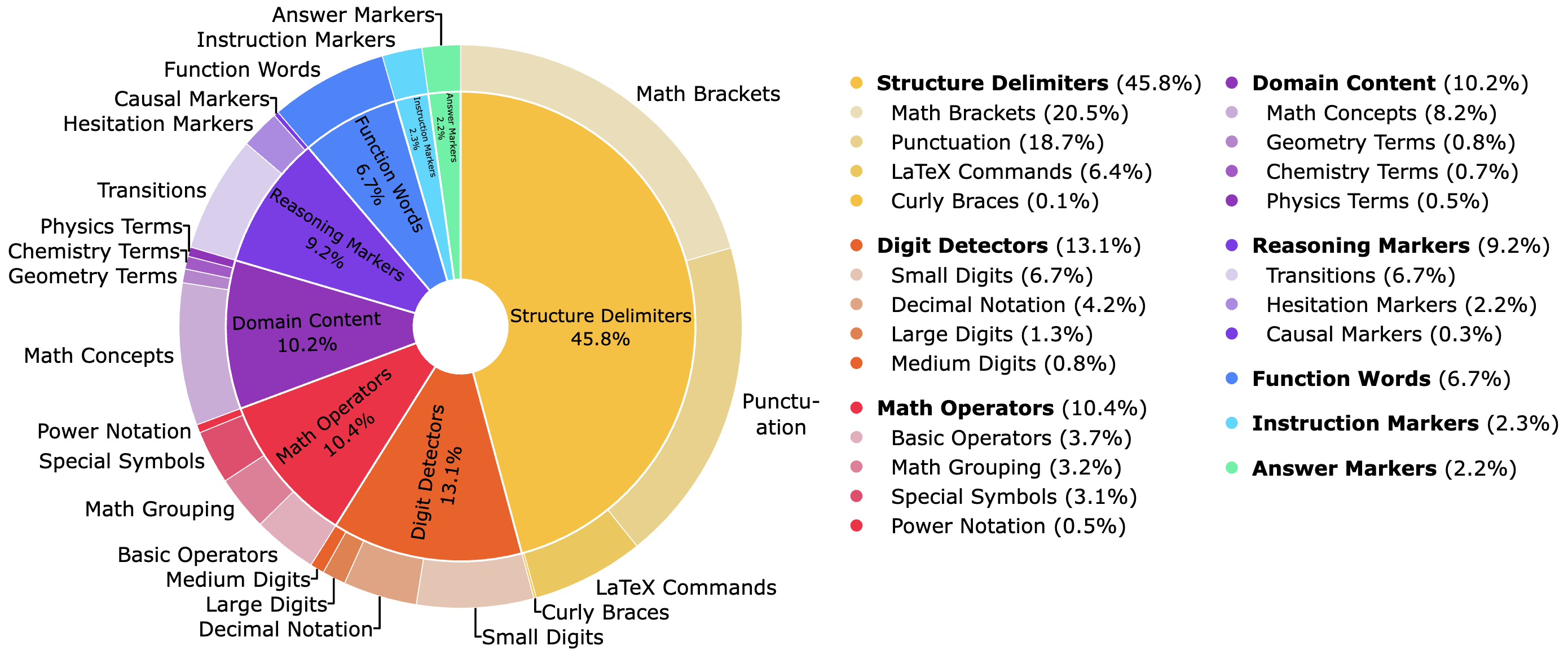}
    \caption{Overview of feature categories learned by an SAE trained on LoRA activation states. Percentages indicate relative feature activation densities. Categories and subcategories were generated by prompting an LLM with feature descriptions.}
    \label{fig:pie_chart}
\end{figure}

\paragraph{Cross-layer SAE over entire adapter state}
To gain a more fine-grained and comprehensive view of features encoded by our LoRA, we train a cross-layer sparse autoencoder (SAE) on the entire 448-dimensional LoRA activation state. We use a batch-top-k SAE with k=16, and an expansion factor of 8 \citep{bussmann2024batchtopk,minder2025crosscoder-sparsity}. After filtering for dead latents, the trained SAE contains roughly 2000 features. We run the same automated interpretation and classification pipeline on max-activating examples for each feature in this SAE, and then further categorize each feature with LLM-generated subcategories. We then visualize relative feature prevalence by computing category-wise activation densities, which is the normalized sum of feature activations within each category over a subset of training data, and plot these in figure \ref{fig:pie_chart}. We note that this categorization methodology is subjective and sensitive to prompt verbiage, and provide this visualization only as a tool to develop high-level intuitions about the adapter's representational budget across layers and components. Broadly, we observe that features tend to concentrate on mathematical operators and syntax, numbers, formatting tokens, and reasoning control-flow. Our LLM-based interpretability classification identifies 62\% of SAE features as ``cleanly monosemantic'', up from 22\% of LoRA features, with an additional 22\% of SAE features classified as ``broad but consistent''. For a more direct look into our SAE, a selection of individual SAE feature activations and interpretations are provided in appendix \ref{sae_appendix}.

\subsection{Ablation Study}
\label{ablation_study}

\begin{figure}[h]
    \centering
    \includegraphics[width=1.0\linewidth]{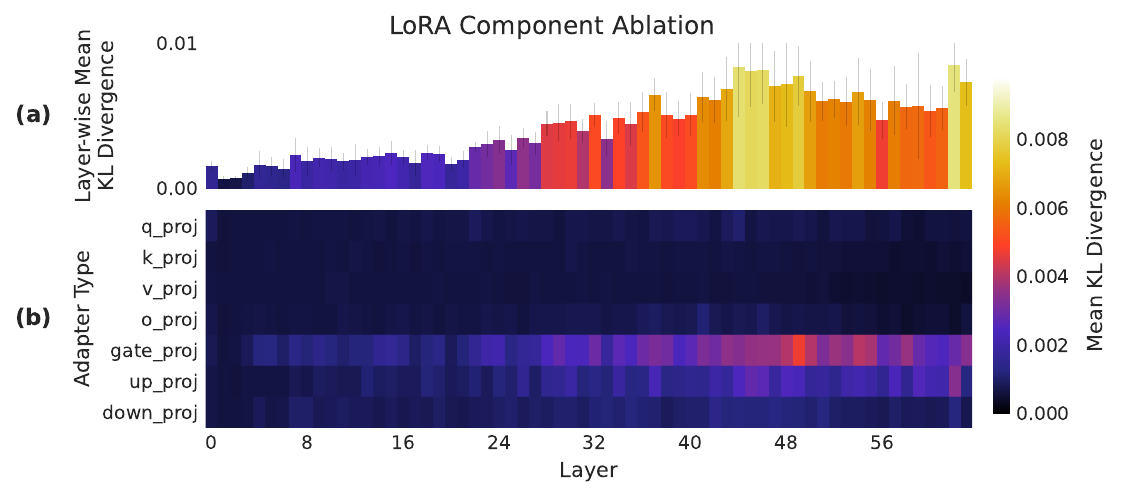}
    \caption{Effect of ablating individual LoRA components from full adapter. \textbf{(a)} Effect of ablating all adapter components at a given layer on KL divergence. \textbf{(b)} Effect of ablating each adapter component individually. Ablation at later layers tends to have a significantly greater effect on the model's output distribution compared to earlier layers. MLP adapters, especially those trained on \texttt{gate\_proj} matrices tend to have the strongest effect.}
    \label{fig:kl_ablation}
\end{figure}

To identify which LoRA components most affect downstream performance, we conduct two ablation experiments. First, we zero out individual components and layers to measure their impact on output KL divergence relative to the unmodified LoRA. Second, we ablate all MLP components simultaneously, and separately all attention components, to assess benchmark performance impact.

Individual component ablation (Figure \ref{fig:kl_ablation}) reveals that mid-to-late layers (particularly 44, 45, 46, and 62) have the greatest effect on downstream KL. MLP components show significantly larger impact than attention components, with \texttt{gate\_proj} having the strongest average effect.

Simultaneous ablation results (Table \ref{tab:lora_ablation_rel_to_lora_plus_opt}) confirm MLP adapters' greater contribution: removing all attention adapters decreases performance but still outperforms the base model, while removing all MLP adapters causes severe degradation, underperforming the base model on one of three tasks and showing poor performance on the others.

\section{Conclusion and Outlook}
\paragraph{Limitations} This work includes several limitations. First, we train our LoRA using a sample-efficient dataset with roughly 10 million tokens. While significant performance can be elicited using this dataset \cite{muennighoff2025s1}, it is likely that our trained models do not contain all of the circuits which are learned by models trained on larger datasets. Additionally, we attempted to prove that extracted LoRA directions have a causal effect on model outputs when used for steering, but found inconclusive results. Anecdotally, steering experiments required very large steering magnitudes to have noticable effects (in excess of 50x normal activation magnitudes), at which point the model would have a high propensity for backtracking (outputting the \texttt{Wait,} token, among others). More investigation of this phenomenon is required to make claims about what is going on. We also heavily use LLM-based autointerpretation and autocategorization methods in our analysis. These methods are useful for painting a general picture of model mechanisms, but we suspect they often fail to uncover the ``true'' role of extracted features. Finally, we only study one model, \texttt{Qwen-2.5-32B-Instruct}, and evaluate this model mostly on math-related reasoning benchmarks.

We demonstrate that reasoning capabilities in large language models can be substantially recovered through minimal parameter modifications, with a rank-1 LoRA recovering 73-90\% of full finetuning performance while having only $\sim0.03\%$ as many trainable parameters. Our analysis reveals that these minimal parameter changes encode interpretable, reasoning-specific signals, with individual adapter directions exhibiting monosemantic properties. Further, we find that sparse autoencoders are useful for extracting additional monosemantic features from LoRA activation states. These findings open new avenues for understanding LLMs through parameter-efficient methods, and suggest that similar targeted approaches could be useful to study other emergent capabilities. Future work could involve focused analysis identifying specific circuits that LoRAs interface with, which we hope could illuminate the core computational mechanisms underlying reasoning behavior in language models.

\section{Acknowledgments}
This work was completed during the ML Alignment and Theory (MATS) 8.0 program. JW thanks Elizabeth Donoway for useful discussions at the beginning of this project, and Bryce Woodworth for support throughout.

\bibliographystyle{plainnat}
\bibliography{references}

@article{wei2022chainofthought,
  title        = {Chain-of-Thought Prompting Elicits Reasoning in Large Language Models},
  author       = {Wei, Jason and Wang, Xuezhi and Schuurmans, Dale and Bosma, Maarten and Xia, Fei and Chi, Ed H. and Le, Quoc V. and Zhou, Denny and others},
  journal      = {arXiv preprint arXiv:2201.11903},
  year         = {2022},
  url          = {https://arxiv.org/abs/2201.11903}
}

@article{kojima2022zeroshot,
  title        = {Large Language Models are Zero-Shot Reasoners},
  author       = {Kojima, Takeshi and Gu, Shixiang Shane and Reid, Machel and Matsuo, Yutaka and Iwasawa, Yusuke},
  journal      = {arXiv preprint arXiv:2205.11916},
  year         = {2022},
  url          = {https://arxiv.org/abs/2205.11916}
}

@techreport{openai2024o1systemcard,
  title        = {OpenAI o1 System Card},
  institution  = {OpenAI},
  year         = {2024},
  month        = {December},
  url          = {https://cdn.openai.com/o1-system-card-20241205.pdf}
}

@article{muennighoff2025s1,
  title        = {s1: Simple Test-time Scaling},
  author       = {Muennighoff, Niklas and Yang, Zitong and Shi, Weijia and Li, Xiang Lisa and Fei-Fei, Li and Hajishirzi, Hannaneh and Zettlemoyer, Luke and Liang, Percy and Cand{\`e}s, Emmanuel and Hashimoto, Tatsunori},
  journal      = {arXiv preprint arXiv:2501.19393},
  year         = {2025},
  url          = {https://arxiv.org/abs/2501.19393}
}

@misc{lindsey2024crosscoders,
  title        = {Sparse Crosscoders for Cross-Layer Features and Model Diffing},
  author       = {Lindsey, Jack and Templeton, Adly and Marcus, Jonathan and Conerly, Thomas and Batson, Joshua and Olah, Christopher},
  howpublished = {\url{https://transformer-circuits.pub/2024/crosscoders/}},
  year         = {2024},
  note         = {Transformer Circuits Thread}
}

@misc{bills2023neurons,
  title        = {Language Models Can Explain Neurons in Language Models},
  author       = {Bills, Steven and Cammarata, Nick and Mossing, Dan and Tillman, Henk and Gao, Leo and Goh, Gabriel and Sutskever, Ilya and Leike, Jan and Wu, Jeff and Saunders, William},
  year         = {2023},
  howpublished = {\url{https://openaipublic.blob.core.windows.net/neuron-explainer/paper/index.html}},
  institution  = {OpenAI}
}

@article{hu2022lora,
  title        = {LoRA: Low-Rank Adaptation of Large Language Models},
  author       = {Hu, Edward J. and Shen, Yelong and Wallis, Phillip and Allen-Zhu, Zeyuan and Li, Yuanzhi and Wang, Shean and Wang, Lu and Chen, Weizhu},
  journal      = {arXiv preprint arXiv:2106.09685},
  year         = {2022},
  url          = {https://arxiv.org/abs/2106.09685}
}

@article{baek2025distilled,
  title         = {Towards Understanding Distilled Reasoning Models: A Representational Approach},
  author        = {Baek, David D. and Tegmark, Max},
  journal       = {arXiv preprint arXiv:2503.03730},
  year          = {2025},
  url           = {https://arxiv.org/abs/2503.03730},
  archivePrefix = {arXiv},
  eprint        = {2503.03730},
  primaryClass  = {cs.LG},
  note          = {ICLR 2025 Building Trust Workshop (paper)}
}

@article{venhoff2025understanding,
  title         = {Understanding Reasoning in Thinking Language Models via Steering Vectors},
  author        = {Venhoff, Constantin and Arcuschin, Iv{\'a}n and Torr, Philip and Conmy, Arthur and Nanda, Neel},
  journal       = {arXiv preprint arXiv:2506.18167},
  year          = {2025},
  url           = {https://arxiv.org/abs/2506.18167},
  doi           = {10.48550/arXiv.2506.18167},
  archivePrefix = {arXiv},
  eprint        = {2506.18167},
  primaryClass  = {cs.LG},
  note          = {ICLR 2025 Workshop on Reasoning and Planning for LLMs (paper)}
}

@article{ward2025repurposes,
  title         = {Reasoning-Finetuning Repurposes Latent Representations in Base Models},
  author        = {Ward, Jake and Lin, Chuqiao and Venhoff, Constantin and Nanda, Neel},
  journal       = {arXiv preprint arXiv:2507.12638},
  year          = {2025},
  url           = {https://arxiv.org/abs/2507.12638},
  doi           = {10.48550/arXiv.2507.12638},
  archivePrefix = {arXiv},
  eprint        = {2507.12638},
  primaryClass  = {cs.LG},
  note          = {ICML 2025 Workshop on Actionable Interpretability}
}

@article{mukherjee2025rlsubnetworks,
  title         = {Reinforcement Learning Finetunes Small Subnetworks in Large Language Models},
  author        = {Mukherjee, Sagnik and Yuan, Lifan and Hakkani-Tur, Dilek and Peng, Hao},
  journal       = {arXiv preprint arXiv:2505.11711},
  year          = {2025},
  url           = {https://arxiv.org/abs/2505.11711},
  doi           = {10.48550/arXiv.2505.11711},
  archivePrefix = {arXiv},
  eprint        = {2505.11711},
  primaryClass  = {cs.LG}
}

@misc{choi2024scaling,
  title        = {Scaling Automatic Neuron Description},
  author       = {Choi, Dami and Huang, Vincent and Meng, Kevin and Johnson, Daniel D. and Steinhardt, Jacob and Schwettmann, Sarah},
  year         = {2024},
  month        = oct,
  howpublished = {\url{https://transluce.org/neuron-descriptions}},
  institution  = {Transluce AI},
  note         = {Published October 23, 2024. Accessed August 22, 2025},
  urldate      = {2025-08-22}
}

@article{bussmann2024batchtopk,
  title         = {BatchTopK Sparse Autoencoders},
  author        = {Bussmann, Bart and Leask, Patrick and Nanda, Neel},
  journal       = {arXiv preprint arXiv:2412.06410},
  year          = {2024},
  url           = {https://arxiv.org/abs/2412.06410},
  doi           = {10.48550/arXiv.2412.06410},
  archivePrefix = {arXiv},
  eprint        = {2412.06410},
  primaryClass  = {cs.LG}
}

@article{minder2025crosscoder-sparsity,
  title         = {Overcoming Sparsity Artifacts in Crosscoders to Interpret Chat-Tuning},
  author        = {Minder, Julian and Dumas, Cl{\'e}ment and Juang, Caden and Chugtai, Bilal and Nanda, Neel},
  journal       = {arXiv preprint arXiv:2504.02922},
  year          = {2025},
  url           = {https://arxiv.org/abs/2504.02922},
  doi           = {10.48550/arXiv.2504.02922},
  archivePrefix = {arXiv},
  eprint        = {2504.02922},
  primaryClass  = {cs.LG}
}


\appendix
\section{Appendix}
\subsection{Model Weights}
Weights for the trained LoRA are available at \url{https://huggingface.co/jnward2/qwen-reasoning-lora}.

\subsection{Selected LoRA Directions}
\label{lora_appendix}
We present a cherry-picked selection of interesting and interpretable LoRA directions, visualized with a dashboard. The full dashboard is available at \url{https://jakeward.page/lora_sae_dashboard.html}. On the left of the dashboard are max-activating examples of the direction over the training dataset. On the right, a full training sample is shown with activations highlighted. Highlight color represents whether the activation is positive or negative.

\begin{figure}[H]
    \centering
    \includegraphics[width=1.0\linewidth]{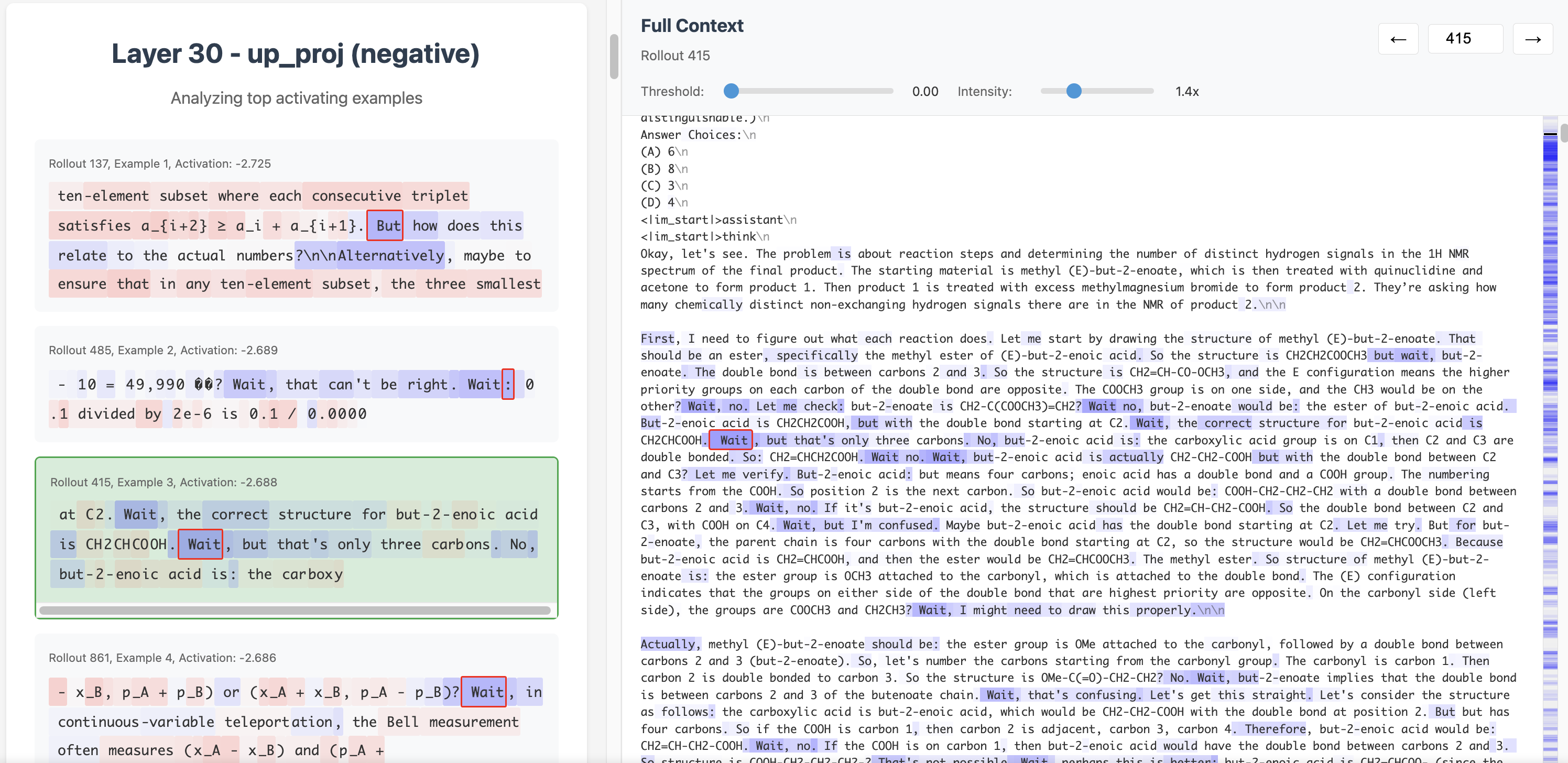}
    \caption{Autointerpretation: \textit{Consistently activates on the token ``Wait'' (and surrounding commas/periods) across examples}}
    \label{fig:placeholder}
\end{figure}

\begin{figure}[H]
    \centering
    \includegraphics[width=1.0\linewidth]{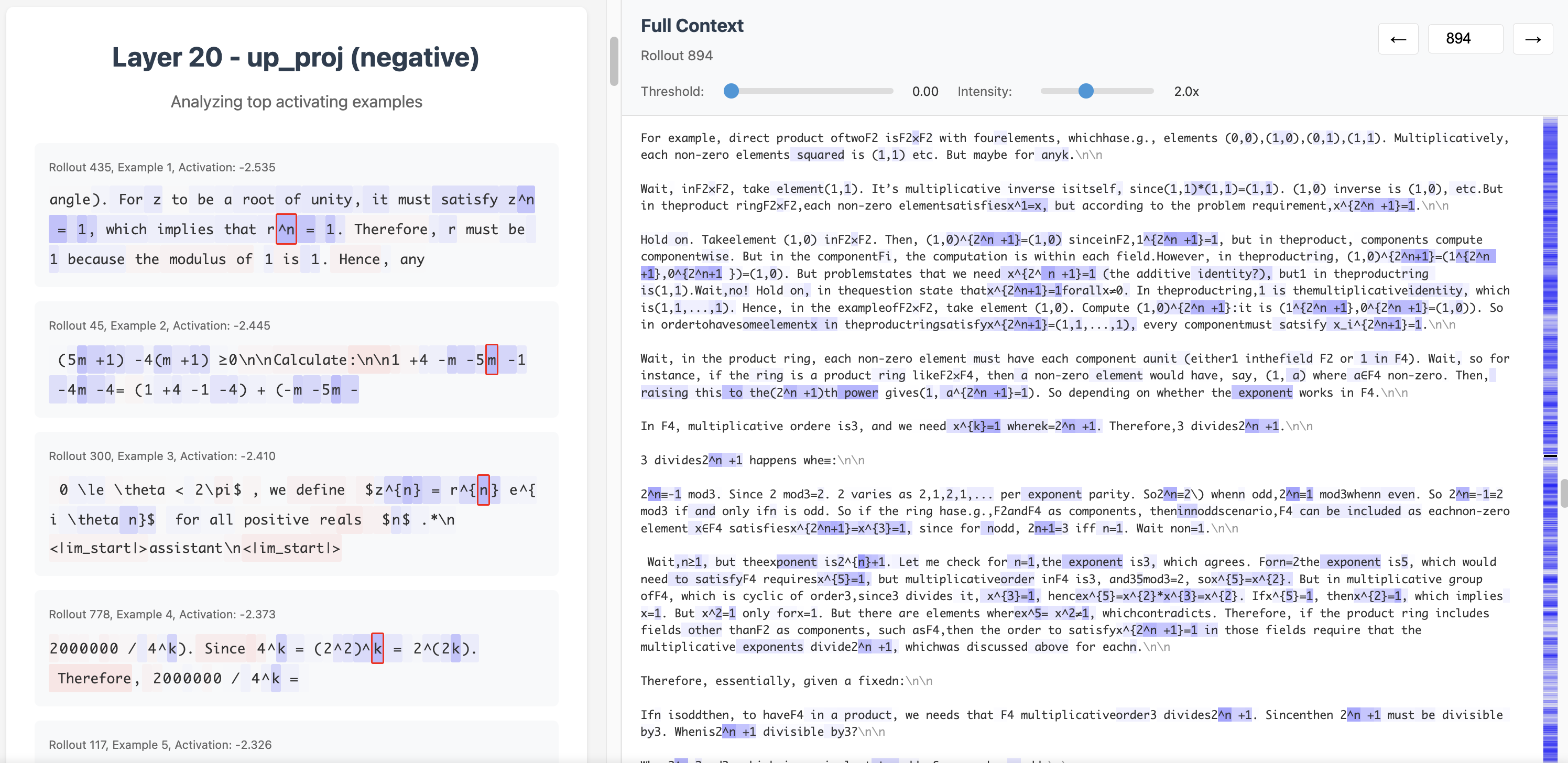}
    \caption{Autointerpretation: \textit{single-letter math variables and exponent tokens}}
    \label{fig:placeholder}
\end{figure}

\begin{figure}[H]
    \centering
    \includegraphics[width=1.0\linewidth]{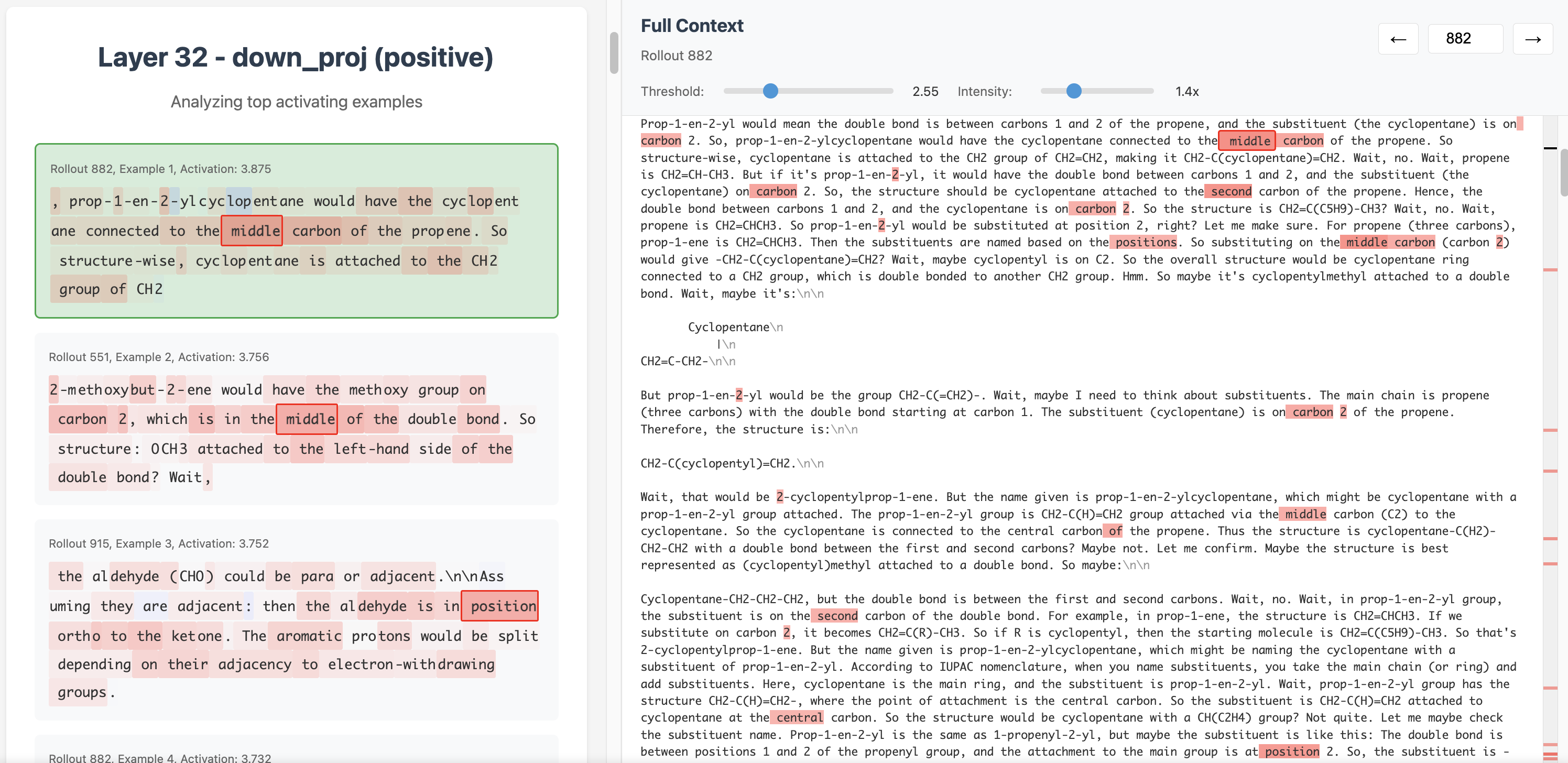}
    \caption{Autointerpretation: \textit{indicates positional numbering in molecules (position, numeral, middle)} Note: a thresholding operation was applied to make visualizing top activations in the full context easier.}
    \label{fig:placeholder}
\end{figure}

\subsection{Selected SAE Features}
\label{sae_appendix}
We additionally present a cherry-picked selection of SAE features. The dashboard screenshots highlight max-activating feature examples over the training dataset.

\begin{figure}[H]
    \centering
    \includegraphics[width=1.0\linewidth]{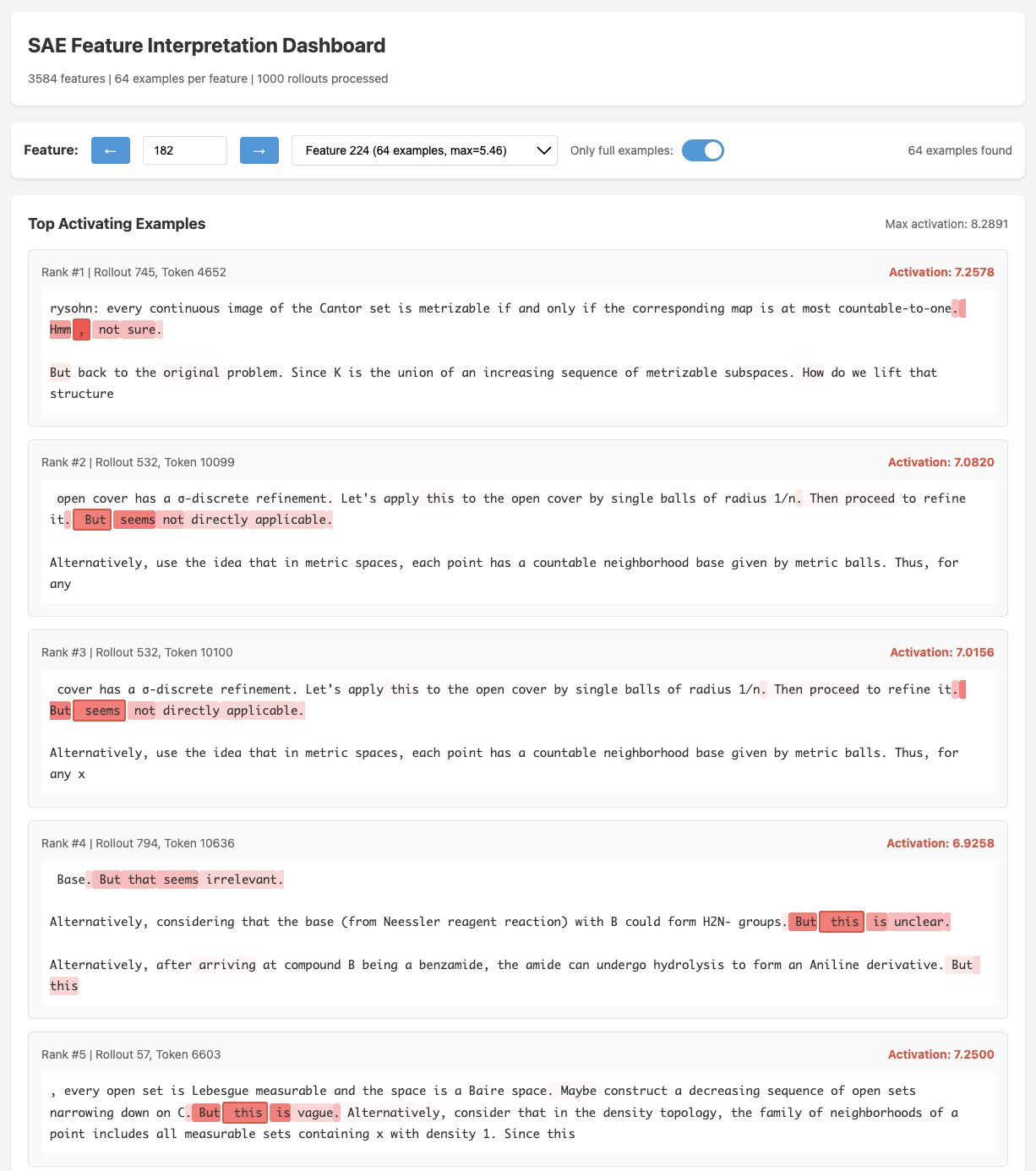}
    \caption{Autointerpretation: \textit{Hesitation markers, notably ``Hmm'', in reflective internal thought.}}
    \label{fig:placeholder}
\end{figure}

\begin{figure}[H]
    \centering
    \includegraphics[width=1.0\linewidth]{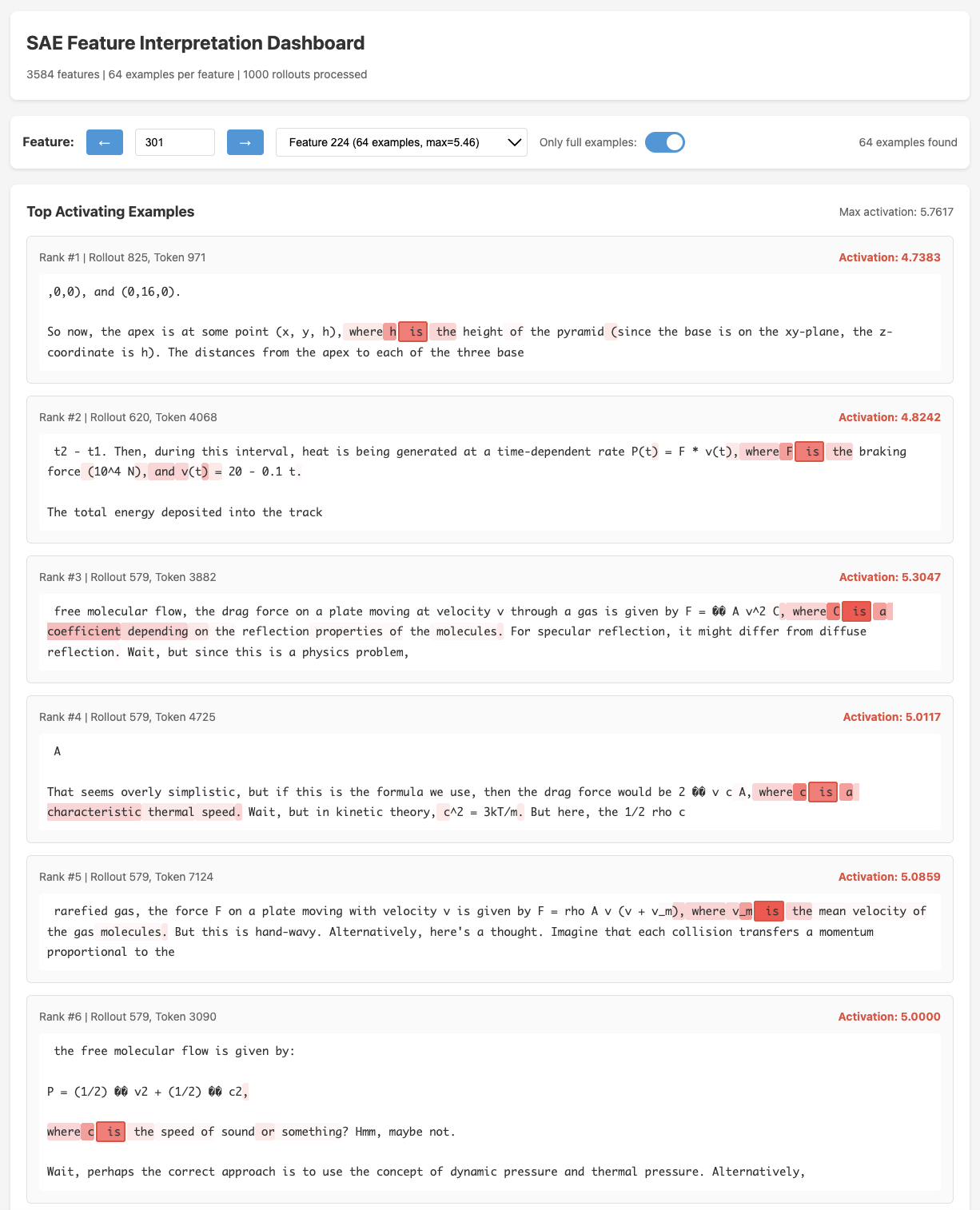}
    \caption{Autointerpretation: \textit{Firing on ``is'' as an equality indicator in mathematical definitions.}}
    \label{fig:placeholder}
\end{figure}

\begin{figure}[H]
    \centering
    \includegraphics[width=1.0\linewidth]{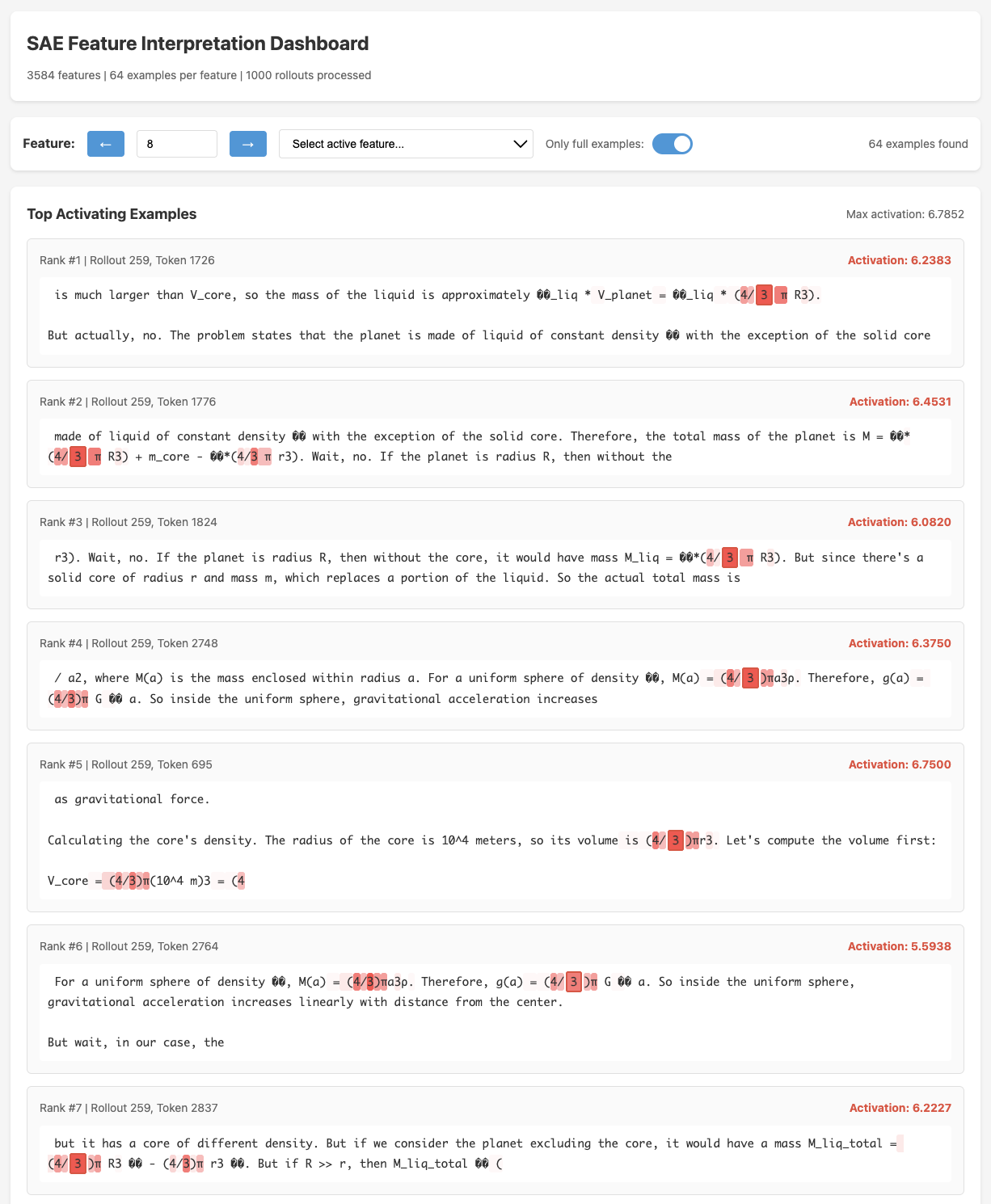}
    \caption{Autointerpretation: \textit{Sphere volume formula tokens, specifically $4/3\pi R^3$ components.}}
    \label{fig:placeholder}
\end{figure}

\begin{figure}[H]
    \centering
    \includegraphics[width=1.0\linewidth]{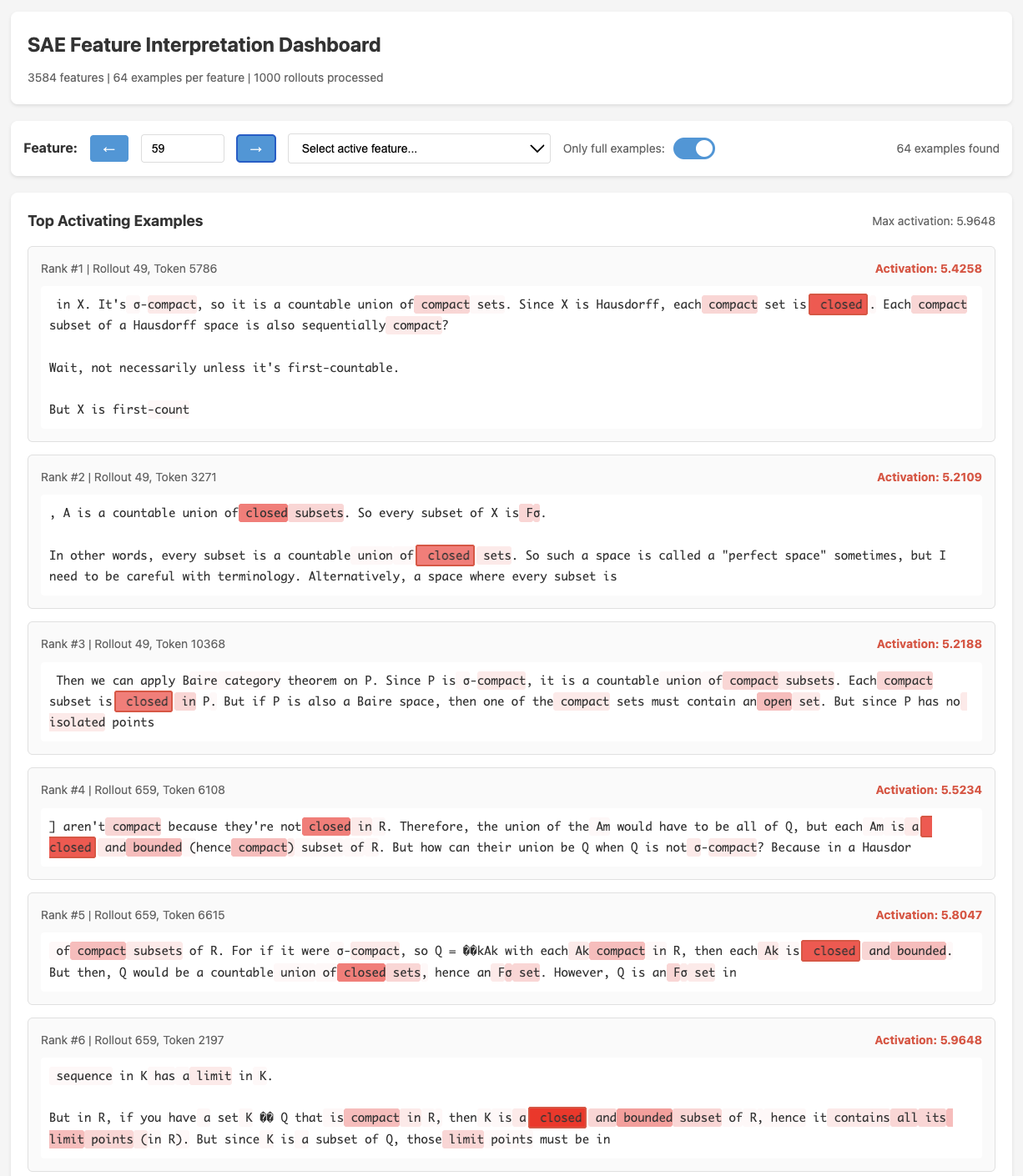}
    \caption{Autointerpretation: \textit{Fires on ``closed'' in mathematical topology contexts.}}
    \label{fig:placeholder}
\end{figure}

\section{LLM Prompts}
\label{llm_prompts}
\subsection{Autointerpretation}
We use \texttt{gpt-5-mini} to automatically generate interpretations for LoRA directions, MLP neurons, and SAE features:
\begin{verbatim}
We're studying neurons in a neural network. Each neuron looks for some particular \
thing in a short document. Look at the parts of the document where the neuron \
activates and describe what it's firing for.

Some activations will be noisy, in these cases you'll have to look for common \
phrases or concepts in the examples.

If a feature always activates for the same token, you should note this in your \
explanation, and also state whether that feature represents that token in some \
specific context. You may need to look at words surrounding activating tokens \
in order to understand why a feature is firing.
Features should have a clear, singular explanation and should be monosemantic. \
If there isn't a clear monosemantic explanation, note this.

Your explanation should not exceed ten words. Don't write complete sentences. \
The neuron might be responding to:
- Individual tokens or specific words
- Phrases or expressions  
- Abstract concepts or behaviors
- Broader context or topics

The activation format shows the full text first, then lists tokens where the neuron \
fired along with their activation strengths (0-10 scale). Higher values mean \
stronger activation.

For example:
cat and mouse ran around the tree. They quickly
tree 7.81
ran 2.30
around 1.01

Look at every example, and then generate an explanation. After generating an \
explanation, assess how monosemantic the feature is.

<neuron_activations>
{activations_str}
</neuron_activations>

Classify the feature's interpretability:
0: The feature is specific, clear, and monosemantic. All given examples clearly \
adhere to the explanation. The explanation is not broad, and the examples are not \
noisy. This feature has a clear and obvious interpretation.
1: The feature may be broad or noisy, but ALL given examples still adhere to the \
generated explanation. The explanation may not be obvious.
2: The feature appears polysemantic. Some examples do not clearly adhere to the \
generated explanation. The explanation does not cleanly explain the given examples.

Respond with JSON in exactly this format:
{{
  "explanation": "your concise explanation in just a few words",
  "classification": <0, 1, or 2>,
  "classification_reasoning": "brief justification for your classification"
}}
\end{verbatim}
\subsection{Autocategorization}
We use \texttt{claude-opus-4.1} to generate feature categories given feature interpretations using the following prompt:

\begin{verbatim}
You will be provided with a list of interpretations of MLP neurons from a \
large language model. Each interpretation describes what a particular learned \
feature detects or responds to during tasks.
Your task is to identify high-level functional role categories that capture what \
these features are doing computationally.

Read through all feature interpretations carefully
Identify natural groupings based on the computational or functional role each \
feature plays
Generate 5-8 high-level categories that capture the major functional roles

Important considerations:
- Focus on computational function (what role the feature plays) rather than \
surface-level similarity
- Categories should be mutually exclusive when possible, though some features \
may have dual roles
- Aim for categories that would generalize across different types of tasks
- Include 3-5 example feature interpretations for each category (copied exactly \
from the list provided)

Output your response as a JSON object with the following structure:
```json
{{
  "categories": [
    {{
      "string_id": "category_identifier_in_snake_case",
      "name": "Human Readable Category Name",
      "definition": "A 1-2 sentence description of what functional role these features serve.",
      "examples": [
        "exact feature interpretation from the list",
        "another feature interpretation from the list",
        "additional examples as needed"
      ]
    }}
  ],
  "summary": "Your overall categorization logic and any notable patterns you observed"
}}
```

Ensure your response contains ONLY the JSON object, with no additional text before or after.

Here are the feature interpretations to categorize:
{feature_list}
\end{verbatim}

Given these categories, we categorize each feature using \texttt{gpt-5-mini} using the following prompt:
\begin{verbatim}
Categorize this neural network feature into ONE of the given categories.

Feature explanation: "{feature.explanation}"

Activation examples (showing what tokens/contexts activate this feature):
{examples_str}

Available categories:
{categories_str}

Based on the feature explanation and activation examples, which category best fits this feature?
Reply with ONLY the category string_id, nothing else.

Your response:
\end{verbatim}

\section{Additional Data}

\begin{table}[h]
\centering
\small
\begin{tabular}{lcccc}
\toprule
\textbf{Task} & \textbf{Full LoRA} & \textbf{Full LoRA (attn-ablated)} & \textbf{Full LoRA (mlp-ablated)}\\
\midrule
AIME'24 (no-figures) & 0.5000 (100.00\%) & 0.3667 (50.02\%) & 0.1333 ($-37.50$\%)\\
MATH500              & 0.9100 (100.00\%) & 0.9000 (86.84\%) & 0.8440 (13.16\%)\\
GPQA-Diamond         & 0.5808 (100.00\%) & 0.5152 (27.83\%) & 0.5051 (16.72\%)\\
\end{tabular}
\caption{Absolute scores with percentages in parentheses, computed relative to the Full Rank-1 LoRA’s recovered gain: $(x-b)/(\ell-b)\times100\%$, where $b$ is the baseline and $\ell$ is the Full LoRA score (so Full LoRA is 100\%).}
\label{tab:lora_ablation_rel_to_lora_plus_opt}
\end{table}


\end{document}